# Multimodal Deep Learning for Diabetic Foot Ulcer Staging Using Integrated RGB and Thermal Imaging

Gülengül Mermer, Mustafa Furkan Aksu, Gözde Özsezer, Şevki Çetinkalp, Orhan Er, Mehmet Kemal Güllü

## 1. Introduction

Diabetes Mellitus (DM), a chronic disease resulting from impaired insulin metabolism, is a public health issue affecting the entire world. DM affects 11.1% of adults aged 20–79 globally (≈589 million people). This number is expected to reach approximately 853 million (1 in 8 adults) by 2050 (International Diabetes Federation (IDF), 2025). DM caused approximately 3.4 million deaths in 2024 and accounted for a significant share of the global health budget, with healthcare expenditures reaching US$1 trillion (International Diabetes Federation (IDF), 2025). These data highlight the urgency of approaches focused on early diagnosis, monitoring, and prevention of complications to sustainably manage the costs, mortality, and morbidity associated with DM.

Diabetic foot ulcer (DFU), one of the most significant complications of DM, leads to hospital admissions due to hospitalisations and non-traumatic lower limb amputations. DFU places a burden on healthcare systems through disability, loss of quality of life, and high costs (The International Working Group on Diabetic Foot, 2023). The 2023 IWGDF guidelines prioritise risk classification, regular foot examinations, education, and systematic implementation of preventive care; home-based monitoring and integration of early warning mechanisms into clinical workflows to reduce this burden.

The risk of recurrence in DFU is high. Recurrence rates of 25–44 per cent within one year and 50–65 per cent within three to five years have been reported following healing (Armstrong et al., 2023; McDermott et al., 2023). Recurrence increases the likelihood of reinfection, hospitalisation, and amputation along with loss of function .(Armstrong et al., 2023). Therefore, clinical care should focus not only on closing the existing ulcer but also on maintaining remission (Armstrong et al., 2023; McDermott et al., 2023).

In terms of amputation risk, a recent meta-analysis reported a lower limb amputation rate of 31% in individuals with DFU. This demonstrates that ineffective management of recurrence and infection leads to severe clinical outcomes (Luo et al., 2024). Multidisciplinary care and close monitoring significantly reduce major amputation rates. This supports the clinical and economic benefits of early diagnosis and rapid intervention (Armstrong et al., 2023; Luo et al., 2024).

Guidelines particularly emphasise regular foot examinations, increased screening frequency based on risk level, daily self-examination, and early reporting of pre-ulcerative changes to reduce the risk of recurrence and amputation. It is also noted that methods such as skin temperature self-monitoring in high-risk groups may help prevent initial or recurrent plantar ulcers (Bus et al., 2024; The International Working Group on Diabetic Foot, 2023).

The classic approach to DFU treatment relies on regular outpatient visits and self-care education. However, dependence on clinic visits (logistical/economic burden), advanced age, and comorbidities often result in poor self-care compliance and mobility limitations. Although current guidelines recommend regular foot examinations, increased screening frequency based on risk level, and reporting of early warning signs at home, standardising home monitoring in terms of fixed distance, angle, and lighting in a repeatable and user-independent manner is challenging (The International Working Group on Diabetic Foot, 2023)

While telehealth and remote plantar temperature monitoring show promise for DFU, the evidence is complex. Multicentre evaluations of remote temperature monitoring (RTM) programmes using systems such as Podimetrics SmartMat™ have not shown a significant reduction in amputation or hospitalisation from all causes in high-risk individuals (Littman et al., 2023). In contrast, beneficial clinical outcome associations have been reported in those using RTM in the US Veterans Health Administration (VHA). However, findings should be interpreted cautiously in terms of design and confounders (Littman et al., 2024). Although the IWGDF (2023) guideline states that skin temperature monitoring may contribute to preventing ulcer recurrence in high-risk groups (e.g., using a ≥2.2 °C difference between feet as a warning threshold), this approach must be supported by standard in-home capture geometry.

Smartphone-based solutions such as Foot Selfie have facilitated remote monitoring of DFU through daily/weekly image sharing. Acceptability and feasibility were demonstrated in two pilot studies conducted in 2021 and 2023 (Swerdlow et al., 2023). However, as these approaches rely on manual capture and user technique, there is high variability in distance, angle, and lighting. This can increase the margin of error in longitudinal comparisons of ulcer margins, skin colour, and thermal markers. In a similar study, the MyFootCare mobile application for DFU monitoring facilitated wound area tracking. However, measurement noise was observed in captures where hands were not free and standard positioning was not maintained (Ploderer et al., 2023). Recent studies highlight the potential of digital health and telehealth in diabetes care, while pointing to evidence heterogeneity, application/participation barriers, and lack of in-home standardisation as key limitations (Maida et al., 2025; Méndez et al., 2025; Zhuang et al., 2025). This gap points to the need for new in-home platforms that enhance clinically meaningful longitudinal comparisons and RGB and thermal integration.

The digital health ecosystem is rapidly advancing in the detection, classification, and recovery prediction of DFU using image-based deep learning. Recent studies have reported high accuracies for ulcer detection, localisation, and staging using CNN/transformer-based models. Some frameworks have demonstrated sensitivity/specificity and AUC above 90% by increasing clinical interpretability with explainable artificial intelligence (XAI) (Cassidy et al., 2023; Rathore et al., 2025). However, some studies emphasise that generalisability remains a challenge due to reasons such as data heterogeneity, limited external validation, and small/biased datasets (Alkhalefah et al., 2025; Silva et al., 2025).

The DFUCare application demonstrated remote and low-cost analysis by reporting wound localisation (mAP ≈ 0.85) and infection/ischaemia classification (≈ 80–95% accuracy) predictions using YOLOv5s on images captured with a smartphone (Sendilraj et al., 2024). The

STANDUP dataset provided a research infrastructure for the use of RGB and thermal multi-modal images in early DFU detection (Bouallal et al., 2022).

With the rapid proliferation of technology in healthcare, interest in smart mirrors has increased; systems have been developed that enable exercise and movement correction at home without seeing a specialist (Kim et al., 2020; Park et al., 2021). ResolDepMirror (Aziz et al., 2021), which detects emotions and provides recommendations based on depression levels; a talking smart mirror (Silapasuphakornwong & Uehira, 2021) that aims to contribute to the daily mood monitoring of elderly people and Alzheimer's treatment; the iMirror, which automatically detects stress (Rachakonda et al., 2020), the Medical Mirror, which estimates pulse from facial optical signals (Poh et al., 2010), and the Wize Mirror, which predicts cardiometabolic risk/anxiety using facial anthropometry (Andreu et al., 2016). Although there are many smart mirrors in the literature focused on health maintenance and improvement, no smart mirror designed for DFU has been found.

In general, smart mirror models are configured with components such as Raspberry Pi, LCD/LED, two-way acrylic mirror, and voice command/remote input; web-API-based content, biometric authentication, multimedia, and personalised profiles (Aanandhi et al., 2021; Sahana et al., 2021). The DFU smart mirror developed within this framework facilitates patients' lives and helps protect individuals at risk from developing ulcers through thermal imaging.

DFU facilitates coordinated and timely intervention in multidisciplinary care (Aalaa et al., 2021; Sorber & Abularrage, 2021). Remote sensing/monitoring reduces patient and system risk by strengthening active surveillance outside the hospital (Rogers et al., 2023). Swerdlow and colleagues (2021) demonstrated plantar surface 'selfie' monitoring using a smartphone. The proposed system in the prototype developed in this study goes beyond this by offering thermal early diagnosis, one-touch operation, and Wagner staging with an RGB camera.

The aim of the study is to investigate the effect of using RGB and thermal images together on model performance in classifying the stages of diabetic foot ulcers and to reveal the improvements that a multi-modality-based deep learning approach will provide. To this end, a Raspberry Pi-based image acquisition system that simultaneously records RGB and thermal images was developed to collect the data to be used in the study. Training sets containing only RGB, only thermal, and 4-channel images consisting of a combination of RGB+thermal images were created with the collected data. Pre-trained CNN models are retrained with these three data sets, and a performance comparison is carried out.

## 2. Method

In this study, three distinct training sets were developed for the classification of DFU stages: RGB, Thermal, and composite RGB-Thermal dataset combining these two modalities. The study investigates the complementary efficacy of these modalities; while RGB images capture essential colour and textural features, thermal images provide critical data on heat anomalies associated with inflammation and infection on the wound surface. Transfer learning was applied to these datasets using various pre-trained Convolutional Neural Network (CNN) architectures.

The methodological framework comprises four main components: (i) image acquisition system, (ii) dataset creation, (iii) image pre-processing, (iv) training and evaluation of CNN models.

## 2.1. Image Acquisition System

A Raspberry Pi-based portable imaging system was designed for the acquisition of DFU images. The system integrates an RGB camera module and software that enables the simultaneous triggering of a thermal camera module. The complete system comprises the following components:

1. Rasbperry Pi 4B 8GB Ram
2. Raspberry Pi Camera Module V2
3. Flir Lepton 3.5 Thermal Camera Module
4. Raspberry Pi 7-inch Touchscreen
5. Custom outer casing fabricated using a 3D printer

The Raspberry Pi Camera Module V2 equipped with an 8 Megapixel CMOS sensor with a native resolution of 3240x2464 pixels and is connected to the Raspberry Pi via the Camera Serial Interface (CSI). RGB images are captured at full resolution to preserve detailed textural and colour information of the diabetic foot region. For thermal imaging, a Flir Lepton 3.5 thermal camera module with a spatial resolution of 160x120 pixels is employed and interfaced With the Raspberry Pi via a USB interface.

The enclosure housing the entire system was fabricated from a PLA filament using a 3D printer. This enclosure ensures stable positioning of the RGB and thermal cameras in a side-by-side configuration, allowing the two sensors to share an approximately overlapping Field of View (FoV). A touchscreen mounted on the front of the prototype provides direct interactions with the system. The graphical user interface (GUI), developed specifically for data acquisition, is operated through this display. An image of the developed prototype is presented in Figure 1 and Figure 2. Real Images of the Produced PrototypeFigure 2.

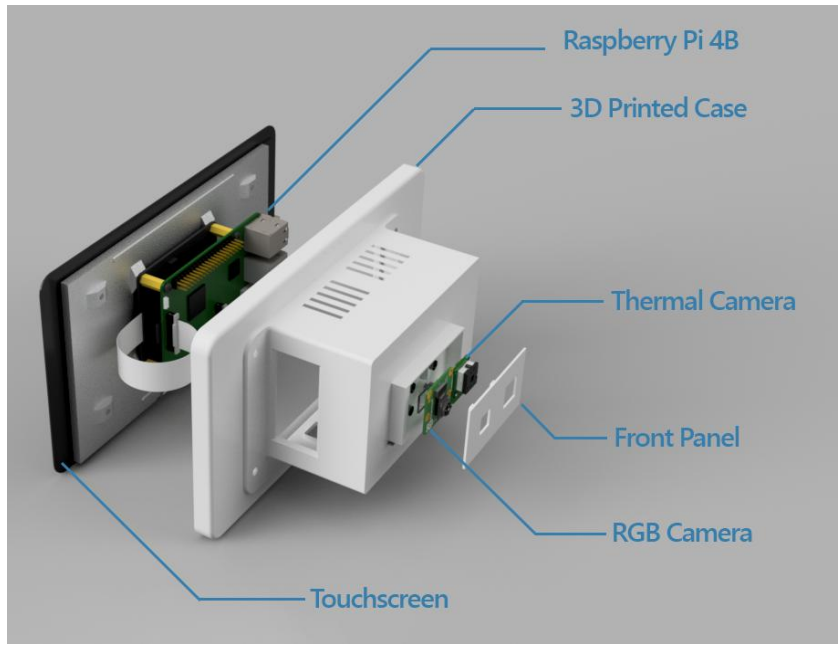

**Figure 1.** Rendered Image of the Produced Prototype

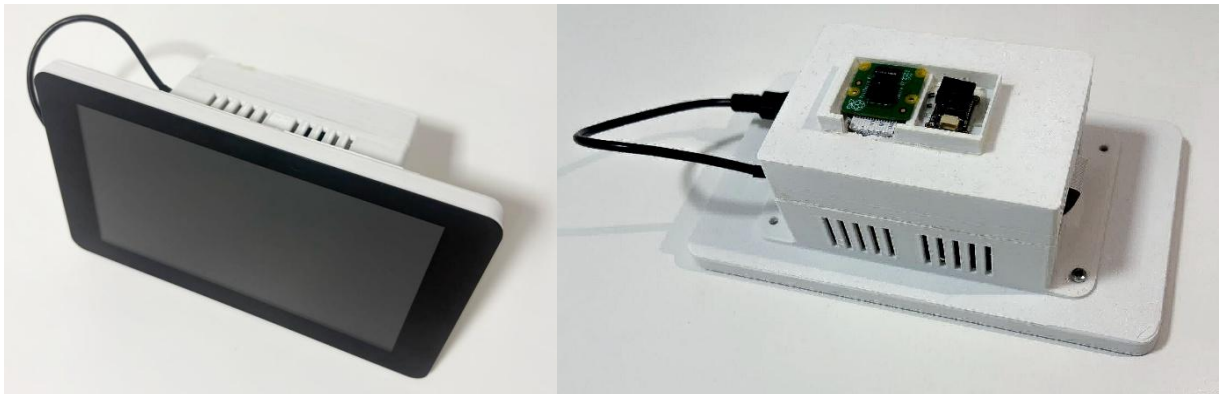

**Figure 2.** Real Images of the Produced Prototype

The designed GUI displays thermal and RGB images simultaneously. Upon user interaction via the designated control button, the RGB image, the colour-mapped thermal image, and the raw thermal image are saved to a directory with a unique timestamp. The RGB and colour-mapped thermal images are stored in JPEG (.jpg) format, whereas the raw thermal data, represented as a temperature matrix, are stored in TIFF (.tiff) format.

### 2.2. Image Pre-processing

Following the creation of the dataset, specific image pre-processing methods were applied to prepare the data for CNN training. RGB images were resized to a standard dimension compatible with the network architectures. For the thermal modality, temperature conversions were performed from the .tiff file containing the raw temperature data. Temperature conversions were performed according to the formula found in the Lepton 3.5 information booklet. The module presents raw temperature data in Kelvin at a resolution of 0.01 and a depth of 14 bits. The conversion from raw pixel data to degrees Celsius is calculated using the following formula.

$$\text{Temperature (°C)} = \frac{\text{Pixel Value}}{100} - 273.15K \qquad (1)$$

To ensure that CNN models focus on the feet in thermal images and to segment the foot area, the obtained temperature matrix was filtered according to the body temperature range. The filtering was performed adaptively, considering the ambient temperature. The corresponding RGB and thermal image pairs in the dataset are shown in Figure 3.

To ensure that CNN models focus on the foot region in thermal images and to segment the medical regions of interest, the obtained temperature matrix was filtered based on the physiological body temperature range. Instead of employing a fixed temperature interval, a dynamic windowing approach was implemented to adapt to ambient temperature fluctuations and sensor calibration shifts. For each image, mean pixel value was calculated and converted into a temperature using Equation 1. If the mean temperature was within the 30 °C - 45 °C range, the image was normalized within these bounds. However, if the mean temperature fell below 30 °C the lower and upper thresholds were iteratively adjusted downwards to preserve thermal contrast. This adaptive filtering ensures that the foot area is consistently highlighted regardless of external thermal noise. The corresponding RGB and thermal image pairs in the dataset, processed through this adaptive normalization, are shown in Figure 3.

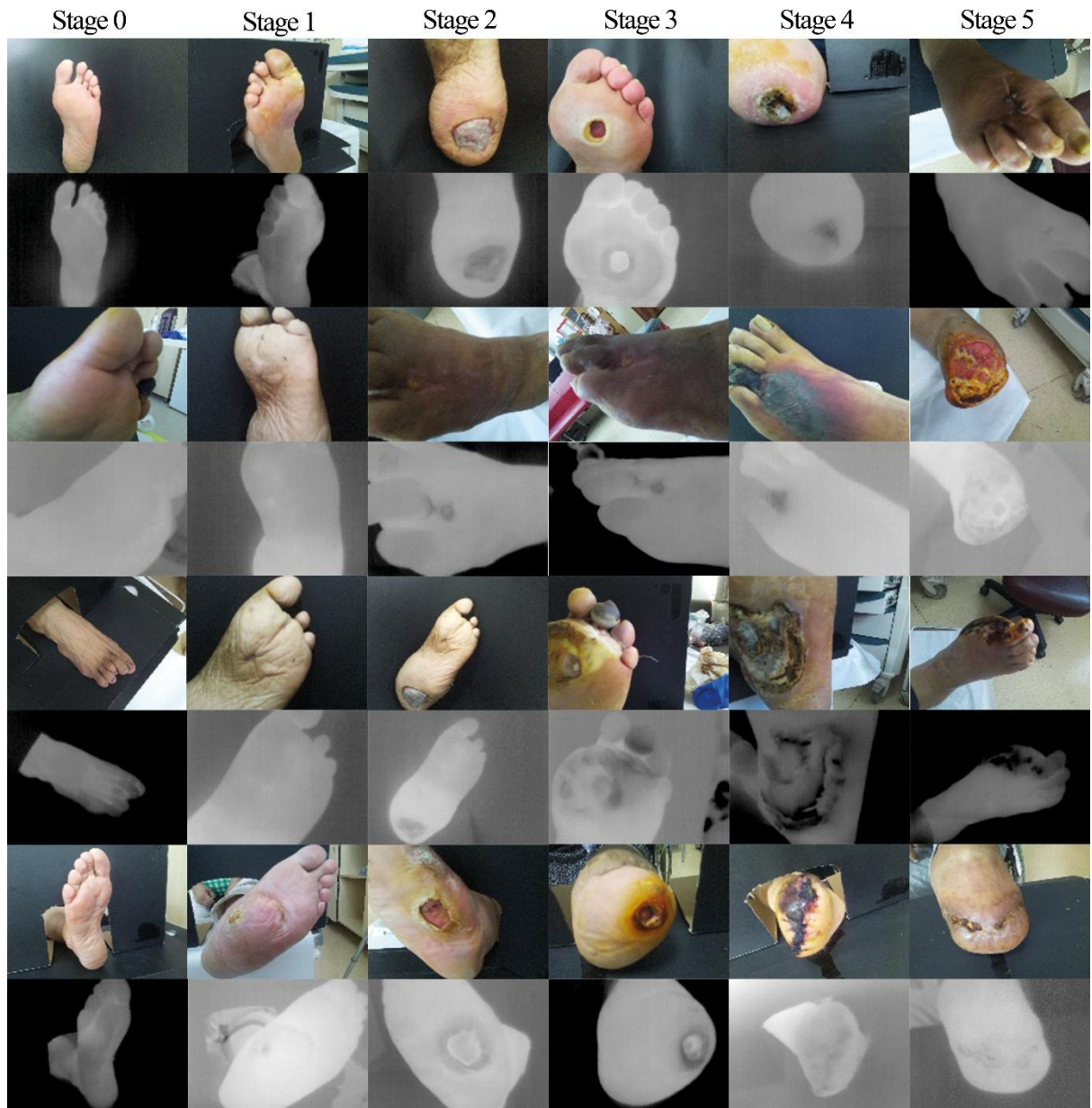

**Figure 3.** RGB and Thermal Image Pairs

## 2.3. Dataset Creation

Data collection was conducted in a hospital setting following ethical approval and the acquisition of informed consent. During the image collection process, three types of images were simultaneously acquired for each patient. The dataset comprises 1205 distinct samples, for which RGB, color-mapped thermal, and raw thermal data were captured simultaneously. These samples were annotated by a clinical expert under six different classes representing different stages of the disease. RGB, thermal, and raw thermal images are available for each sample.

In this study, three distinct training datasets were established to evaluate the efficacy of different modalities: RGB-only, Thermal-only, and composite RGB+Thermal set. For the multimodal configuration, the thermal image was integrated as a fourth channel alongside the RGB input and converted into a unified tensor. The distribution of the number of images for each class in the three training sets is given in Table 1. Because some thermal images exhibited distortions, they were excluded from both the Thermal and RGB+Thermal training sets.

**Table 1.** Number of Images in Training Sets

|  | Grade 0 | Grade 1 | Grade 2 | Grade 3 | Grade 4 | Grade 5 |
|---|---|---|---|---|---|---|
| RGB | 150 | 106 | 496 | 226 | 184 | 43 |
| Thermal | 134 | 84 | 456 | 214 | 179 | 41 |
| RGB+Thermal | 134 | 84 | 456 | 214 | 179 | 41 |

## 2.4. Model Training

To mitigate overfitting and enhance model generalization, data augmentation was applied. These include geometric transformations cropping, resizing, random horizontal flip (p=0.3), random rotation (10°), random affine as well as photometric adjustments utilizing color jitter (brightness=0.2, contrast=0.2, saturation=0.2, hue=0.1).

To reduce class imbalance within the dataset, a class-weighted loss function was implemented. Weights were calculated inversely proportional to the sample frequency of each class. Consequently, minority classes were assigned higher penalty weights during backpropagation This mechanism ensured that the model optimized for all classes in a balanced manner, preventing bias toward the majority classes.

The transfer learning approach was employed for feature extraction. Features were obtained using five different convolutional neural network (CNN) architectures pre-trained on the ImageNet dataset: ResNet-50, VGG16, DenseNet-121, InceptionV3, and EfficientNetV2. All models were initialized with pre-trained weights. The original classification layers of each network were removed and replaced with a customized classification head tailored to the six classes in this study.

The proposed classification head consists of two fully connected layers containing 1024 and 512 neurons, respectively. Each layer is followed by a ReLU activation function, batch normalization, and a dropout layer with a rate of 50%. The final output layer includes six neurons corresponding to the six target classes. This architecture enables more effective representation of high-level features, while the use of dropout and batch normalization helps reduce overfitting and improve training stability. A schematic overview of the proposed methodology is presented in Figure 4. The number of parameters and inference times of the employed models are reported in Table 2.

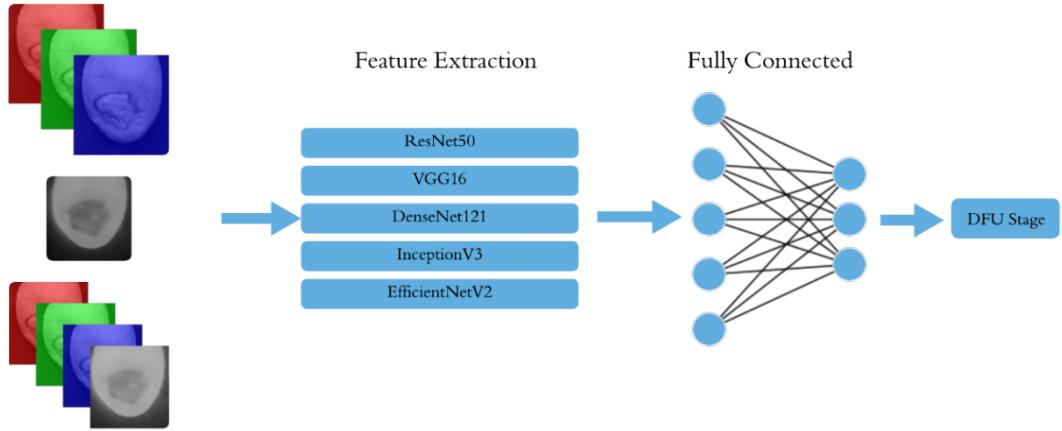

**Figure 4.** Schematic Diagram of Proposed Method

**Table 2.** Parameter Numbers and Inference Times of the Models Used

| Training Set | CNN Model | Number of Parameters (M) | | Inference Time (ms) | | | |
|---|---|---|---|---|---|---|---|
| | | Backbone | Advanced Head | Mean | Max | Min | FPS |
| RGB | DenseNet121 | 6.95 | 1.58 | 6.43 | 6.03 | 7.41 | 155.45 |
| RGB | EfficientNetV2 | 20.17 | 1.84 | 6.94 | 6.44 | 8.97 | 144.07 |
| RGB | InceptionV3 | 21.78 | 5.58 | 4.89 | 4.69 | 5.79 | 204.32 |
| RGB | ResNet50 | 23.50 | 2.62 | 3.29 | 3.08 | 3.81 | 303.94 |
| RGB | VGG16 | 134.26 | 4.72 | 6.02 | 5.87 | 6.72 | 166.02 |
| Thermal | DenseNet121 | 6.95 | 1.58 | 6.35 | 6.03 | 6.98 | 157.45 |
| Thermal | EfficientNetV2 | 20.17 | 1.84 | 6.82 | 6.44 | 7.65 | 146.61 |
| Thermal | InceptionV3 | 21.78 | 5.58 | 4.89 | 4.69 | 5.56 | 204.69 |
| Thermal | ResNet50 | 23.50 | 2.62 | 3.20 | 3.07 | 3.81 | 312.27 |
| Thermal | VGG16 | 134.26 | 4.72 | 6.02 | 5.87 | 7.14 | 166.07 |
| RGB+Thermal | DenseNet121 | 6.95 | 1.58 | 6.34 | 6.08 | 10.41 | 157.70 |
| RGB+Thermal | EfficientNetV2 | 20.17 | 1.84 | 6.96 | 6.61 | 8.54 | 143.67 |
| RGB+Thermal | InceptionV3 | 21.78 | 5.58 | 5.18 | 4.70 | 9.72 | 192.92 |
| RGB+Thermal | ResNet50 | 23.51 | 2.62 | 3.29 | 3.12 | 3.95 | 304.34 |
| RGB+Thermal | VGG16 | 134.28 | 4.72 | 6.01 | 5.89 | 6.75 | 166.34 |

## 3. Experimental Setup and Results

All experiments were conducted using the Python programming language on a workstation equipped with a Ryzen 9 9950X 4.3 GHz CPU, an NVIDIA RTX 3060 GPU, and 32 GB of RAM. The deep learning models were implemented and trained using the PyTorch framework. To reliably assess the generalization performance of the models, a 5-fold stratified cross-validation strategy was employed. Before proceeding with the cross-validation process, 15% of the dataset was set aside as a separate test set and was not included in the training phase. This test set remained separate from all training and validation processes and was used solely to evaluate the final performance of the models trained across the folds. In the five-fold cross-validation scheme, one-fold was used for validation while the remaining four folds were used for training. The results were analyzed in detail using multiple evaluation metrics, including accuracy, precision, recall, F1-score, specificity, and sensitivity. Given the imbalanced class distribution of the dataset, the Matthews Correlation Coefficient (MCC) was additionally employed to provide a more reliable performance assessment. Performance metrics obtained from training on RGB, Thermal, and RGB+Thermal datasets using five different CNN models are presented in Table 3, Table 4 and Table 5. The reported metrics represents the average values across all folds.

**Table 3.** Results of Training with RGB Training Set

| Model | Precision (%) | Recall (%) | F1-Score (%) | Sensitivity (%) | Specificity (%) | MCC (%) | Overall Accuracy (%) |
|---|---|---|---|---|---|---|---|
| DenseNet121 | 90.05 | 84.54 | 86.56 | 84.54 | 97.34 | 85.43 | 89.33 |
| EfficientNetV2 | 88.9 | **86.53** | **87.52** | **86.53** | 97.2 | 84.14 | 87.67 |
| InceptionV3 | 86.41 | 81.45 | 83.09 | 81.45 | 96.57 | 80.57 | 85.11 |
| ResNet50 | 85.49 | 85.48 | 84.83 | 85.48 | 97.28 | 84.22 | 88.00 |
| VGG16 | **90.92** | 84.91 | 86.52 | 84.91 | **97.39** | **86.33** | **90.22** |

**Table 4.** Results of Training with Thermal Training Set

| Model | Precision (%) | Recall (%) | F1-Score (%) | Sensitivity (%) | Specificity (%) | MCC (%) | Overall Accuracy (%) |
|---|---|---|---|---|---|---|---|
| DenseNet121 | 88.27 | 87.24 | 87.28 | 87.24 | 97.05 | 83.64 | 87.71 |
| EfficientNetV2 | 87.35 | 83.77 | 85.11 | 83.77 | 96.55 | 81.08 | 85.78 |
| InceptionV3 | 89.68 | 83.71 | 85.91 | 83.71 | 96.92 | 82.91 | 87.35 |
| ResNet50 | 76.71 | 74.57 | 74.74 | 74.57 | 95.77 | 75.18 | 81.45 |
| VGG16 | **91.66** | **87.87** | **89.44** | **87.87** | **97.84** | **88.04** | **90.96** |

**Table 5.** Results of Training with RGB+Thermal Training Set

| Model | Precision (%) | Recall (%) | F1-Score (%) | Sensitivity (%) | Specificity (%) | MCC (%) | Overall Accuracy (%) |
|---|---|---|---|---|---|---|---|
| DenseNet121 | 91.02 | **93.08** | 91.83 | **93.08** | 97.89 | 87.95 | 91.08 |
| EfficientNetV2 | 89.98 | 87.5 | 88.18 | 87.50 | 97.99 | 88.19 | 91.33 |
| InceptionV3 | 89.73 | 89.25 | 89.18 | 89.25 | 97.52 | 86.04 | 89.40 |
| ResNet50 | 85.33 | 86.56 | 84.87 | 86.56 | 97.51 | 85.59 | 89.76 |
| VGG16 | **94.00** | 91.68 | **92.53** | 91.68 | **98.4** | **91.03** | **93.25** |

An examination of the tables compares the performance metrics obtained across three different data types (RGB, Thermal, RGB+Thermal) and five different CNN architectures. Overall, models trained with the RGB+Thermal dataset achieve the highest values in terms of overall accuracy as well as other metrics Among the evaluated architectures, the VGG16 model demonstrates the best performance consistently across all data types.

In particular, the VGG16 model achieved high accuracy even when trained on RGB and Thermal data individually, while the combination of these two modalities (RGB+Thermal) led to a significant improvement in both overall accuracy and MCC. These results demonstrate that multimodal learning provides a richer and more discriminative representation than single-modality approaches for diabetic foot ulcer classification.

Figure 5 presents a visual comparison of the overall accuracy values obtained by all models using the RGB, thermal, and RGB+Thermal datasets. The graph clearly confirms the trend observed in the numerical results. While accuracy remains lower across all models when using the thermal dataset alone, performance improves significantly when thermal data is combined with RGB inputs. VGG16, DenseNet121, and EfficientNetV2 stand out as the architectures most responsive to multimodal inputs. This visual summary highlights the varying sensitivity of different models to input data types and further supports the superiority of the RGB+Thermal approach.

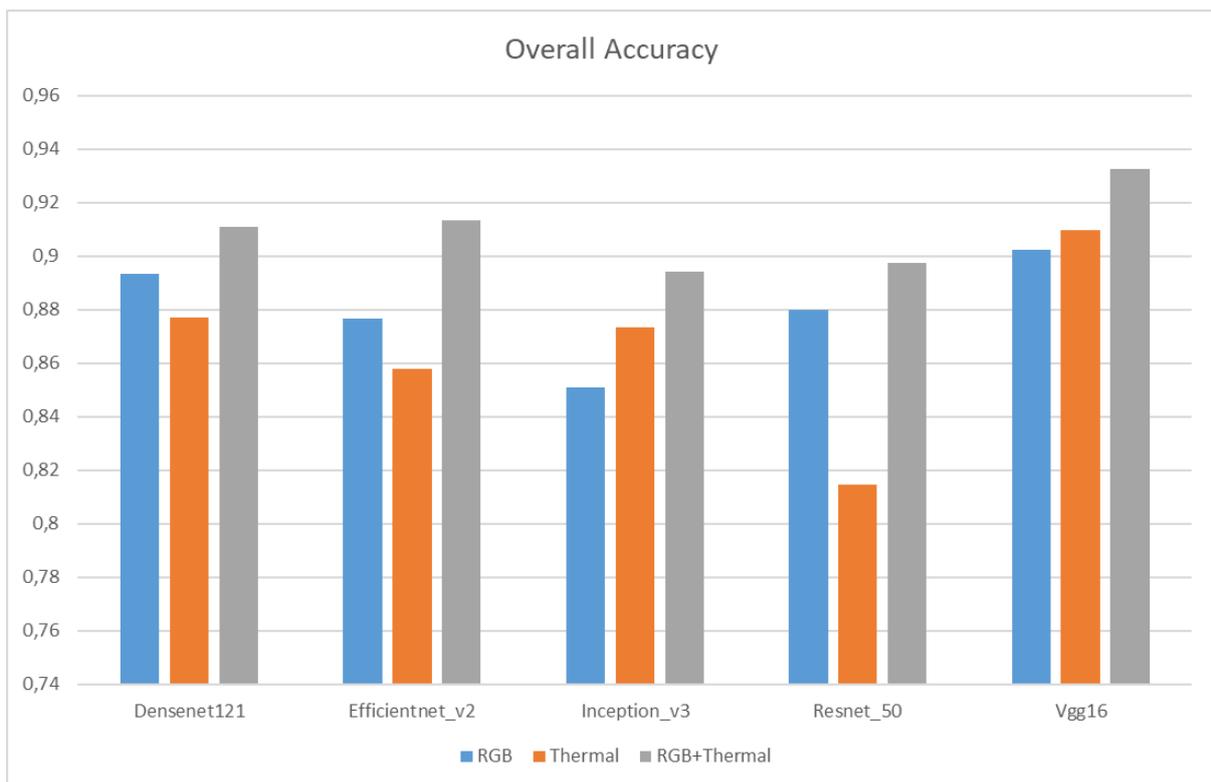

**Figure 5.** Comparison of Overall Accuracy Values of Five CNN Models Trained Using RGB, Thermal And RGB+Thermal Training Sets.

The comparisons presented in the Table 3-5 and Figure 5 demonstrate that the multimodal approach outperforms the single-modality approaches and that the VGG16 model is the most robust architecture evaluated in this study, particularly when trained on the RGB+Thermal dataset. Therefore, further analyses were conducted exclusively on this model. Class-wise performance results for each fold in the 5-fold cross-validation are reported in Table 6.

**Table 6.** Fold-Based Performance Results of the VGG16 Model Trained with RGB+Thermal Training Set

|  | Precision (%) | Recall (%) | F1-Score (%) | Sensitivity (%) | Specificity (%) | MCC (%) | Accuracy (%) | Support |
|---|---|---|---|---|---|---|---|---|
| **Fold 1** | 95.89 | 91.97 | 93.61 | 91.97 | 98.51 | 91.98 | 93.98 | 166 |
| **Fold 2** | 92.82 | 88.45 | 90.05 | 88.45 | 97.75 | 87.23 | 90.36 | 166 |
| **Fold 3** | 93.42 | 94.95 | 93.95 | 94.95 | 98.75 | 92.8 | 94.58 | 166 |
| **Fold 4** | 94.6 | 91.38 | 92.71 | 91.38 | 98.45 | 91.18 | 93.37 | 166 |
| **Fold 5** | 93.26 | 91.65 | 92.35 | 91.65 | 98.53 | 91.97 | 93.98 | 166 |

The fold-based analysis indicates that the RGB+Thermal VGG16 model exhibits high and consistent performance across all data splits. Accuracy, F1-score, and MCC values remain within narrow ranges across all folds, demonstrating strong generalization capability and stable classification performance despite class imbalance. In particular, the 94.58% accuracy and 93.95% F1-score achieved in Fold 3 correspond to the split in which the model most effectively exploits multimodal information. Overall, VGG16 demonstrates high sensitivity, specificity, and MCC values across all folds, confirming that the multimodal approach provides robust and reproducible performance for DFU classification.

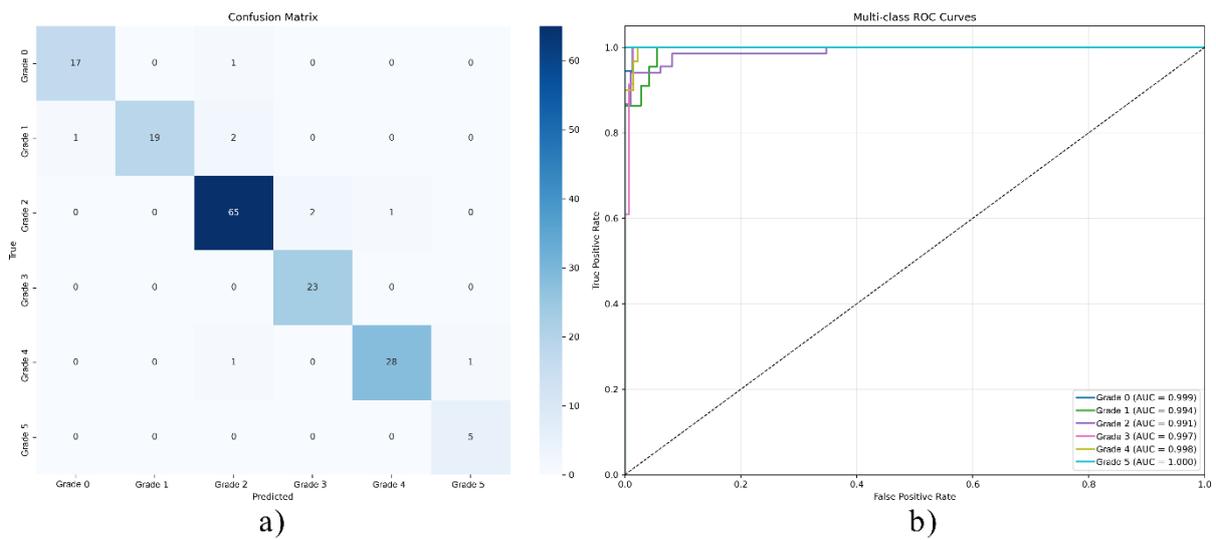

**Figure 6.** Confusion Matrix and ROC Curve of the VGG16 Model Trained with RGB+Thermal Dataset, a) Complexity Matrix, b) ROC Curve

The confusion matrix and ROC curve corresponding to the best-performing fold are presented in Figure 6. The model achieves high performance in Grade 2 (65 correct) and Grade 4 (28 correct). The presence of pronounced textural distortion and significant thermal anomalies in these classes facilitates accurate discrimination. A limited number of misclassifications are observed in the early stages such particularly in Grade 0 and Grade 1. At these stages, the absence of clear ulcer formation and minimal thermal contrast occasionally lead to ambiguity between neighboring classes. However, these misclassifications are concentrated to adjacent classes (e.g., Grade 0 → Grade 1), with no mispredictions involving distant classes, demonstrating that the model's decision behaviour remains consistent with clinical progression.

Gradient-weighted Class Activation Mapping (Grad-CAM) was employed to visualize the image regions influencing the decisions of the VGG16 models trained on the three datasets. This analysis indicates that the model predominantly focuses on clinically relevant regions, including the ulcer region, hyperthermic zones, surrounding inflammation, and disruptions in tissue integrity. Figure 7 shows the Grad-CAM visualizations for the VGG16 models corresponding to the three different datasets.

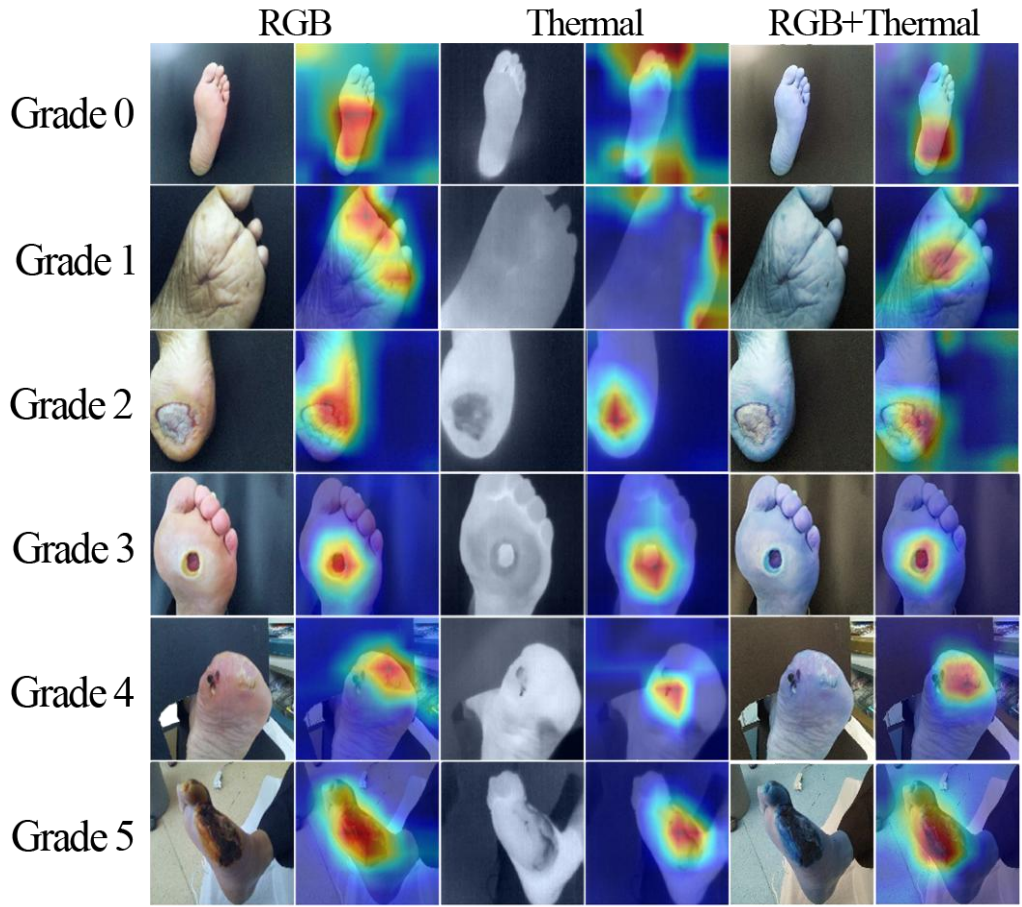

**Figure 7.** Grad-CAM Visualizations Obtained with the VGG16 Models on Different Datasets

The Grad-CAM outputs clearly highlight behavioral differences between the modalities. The model trained solely on the thermal dataset is unable to consistently localize discriminative regions, particularly for Stage 0 and Stage 1 samples. In these early stages, the absence of visible ulceration and the minimal temperature contrast result in lack of distinctive thermal anomalies. Consequently, the model struggles to identify salient regions to support its classification decisions, and the resulting heat maps exhibit weaker and more diffuse activation patterns.

In contrast, the model trained on the RGB dataset generally succeeds in capturing the wound region and the associated textural variations. However, in certain cases, the attention extends beyond the lesion, encompassing a broader area of the foot. This behaviour indicates that reliance solely on morphological information may introduce ambiguity, particularly in lesions with low contrast or unclear boundaries.

In contrast, the RGB+Thermal (multimodal) model, by integrating complementary features from both modalities, exhibits a substantially more stable and clinically meaningful focus pattern. The attention localization issues observed in the thermal-only model, particularly for Stage 0 and Stage 1 samples, are effectively mitigated by the multimodal approach, while the broad and diffuse attentional patterns seen in the RGB-only model are also significantly reduced. As a result, the Grad-CAM maps of the multimodal model show a more compact and well-defined activations that are closely aligned with the lesion areas.

The primary reason for this behaviour is that the thermal modality produces pronounced temperature anomalies in Stage 2 and more advanced lesions. The thermal channel enables the multimodal model to accurately localise the ulcer region, while the RGB channel reinforces the decision-making process by providing complementary textural and structural information. Thus, the RGB+Thermal model effectively integrates the strengths of both modalities, resulting in a more balanced, precise, and clinically meaningful attentional distribution.

## 4. Discussion

The findings of this study reveal that combining RGB and thermal images for diabetic foot ulcer (DFU) stage classification yields significantly higher performance than single modality approaches. Training datasets constructed using RGB, thermal, and RGB+thermal modalities were evaluated across five different CNN architectures. Model performance was assessed using accuracy, F1-Score, and MCC metrics. The highest performance was achieved by the VGG16 model trained on the RGB+thermal modality, demonstrating that multimodal information enhances classification capability. Moreover, the consistent and stable performance observed across all cross-validation folds, despite class imbalance, indicates the strong generalization capacity of the proposed approach.

The complementary physiological information captured by RGB and thermal modalities makes the multimodal approach particularly advantageous. RGB images capture visual cues such as tissue integrity disruption, color variations, and wound boundary structures, whereas thermal images capture temperature differences associated with infected tissue. Spatial temperature differences across the wound region emerge as a key discriminative feature among the classes. However, in the early stages (Grade 0-1), the absence of visible ulceration results in negligible temperature differences, which explains the limited effectiveness of thermal-only classification at these stages. In more advanced stages, thermal images exhibit more pronounced temperature variations, enabling clearer delineation of ulcer boundaries. This effect is particularly evident in Figure 7 for Stages 2 and higher. Consequently, within the multimodal framework, the morphological information provided by RGB images compensates for the limitations of thermal data in early stages, thereby supporting the decision-making process and enabling the model to produce more stable and accurate predictions.

The proposed approach also yields clinically significant implications. The multimodal deep learning-based classification framework developed in this study has the potential to be integrated into home monitoring, telehealth applications, and clinical decision support tools. The simultaneous use of RGB and thermal imaging allows the joint assessment of complementary physiological changes, thereby establishing an infrastructure for the decision support system that can effectively assist Wagner stage classification.

Nevertheless, this study has several limitations. First, the dataset size is relatively limited, and all data were collected at a single center. Additionally, the spatial resolution of the thermal camera is lower than that of the RGB camera, primarily due to the high cost of high-resolution thermal sensors. In order to develop an affordable and practically deployable system, a lower-resolution thermal sensor was therefore preferred, which consequently limits the sensitivity of the thermal modality. Furthermore, differences in optical characteristics and viewing angles between the thermal and RGB cameras prevent precise pixel-level alignment between the two modalities. Another limitation is the class imbalance present in dataset, which may lead to increased variability in classification performance across different stages.

## 5. Conclusion

This study demonstrates that the combined use of RGB and thermal images for diabetic foot ulcer (DFU) stage classification yields higher and more stable performance compared to single-modality approaches. A Raspberry Pi-based image acquisition prototype capable of simultaneously capturing RGB and thermal images was developed, and clinical data were collected using this system. Training datasets were constructed from the collected data using three different modalities and evaluated with five different CNN architectures. Among these modalities, the RGB+thermal configuration-formed by incorporating the thermal image as a fourth channel alongside the RGB image- Achieved the most successful results. This finding confirms that multimodal information substantially enhances the accuracy of DFU stage classification.

The results further demonstrate that the proposed multimodal framework provides a solid foundation for remote DFU monitoring, home-based monitoring systems, telehealth applications, and clinical decision support systems.

Future work may improve the generalizability of the proposed model by incorporating larger and multi-center datasets. In additional, the use of higher-resolution thermal cameras could further enhance the performance of thermal-only and multimodal models by more accurately capturing temperature variations across the wound region. Differences in optical characteristics and viewing angles between the thermal and RGB cameras used in the data acquisition system currently may prevent precise spatial alignment of the two modalities. To address this limitation, automatic image registration techniques may be employed. These improvements are expected to facilitate broader and more reliable deployment of the proposed system in clinical applications.